\documentclass{ecai}
\usepackage{times}
\usepackage{epsfig}
\usepackage{graphicx}
\usepackage{amsmath}
\usepackage{amssymb}
\usepackage{times}
\usepackage{soul}
\usepackage[hidelinks]{hyperref}
\usepackage[utf8]{inputenc}
\usepackage[small]{caption}
\usepackage{graphicx}
\usepackage{amsmath}
\usepackage{booktabs}
\usepackage{multirow}
\usepackage{amsfonts}
\usepackage{multirow}
\usepackage{color}
\usepackage[table]{xcolor}
\definecolor{lightgray}{gray}{0.9}
\definecolor{grey1}{gray}{0.9}

\DeclareMathOperator*{\argmin}{arg\,min}
\title{Ensemble Knowledge Distillation for Learning Improved and Efficient Networks}
\author{Umar Asif \institute{IBM Research Australia, email: umarasif@au1.ibm.com} \and Jianbin Tang \institute{IBM Research Australia, email: jbtang@au1.ibm.com} \and Stefan Harrer\institute{IBM Research Australia, email: sharrer@au1.ibm.com}}
\begin{document}
\maketitle
\begin{abstract}
Ensemble models comprising of deep Convolutional Neural Networks (CNN) have shown significant improvements in model generalization but at the cost of large computation and memory requirements. 
In this paper, we present a framework for learning compact CNN models with improved classification performance and model generalization. For this, we propose a CNN architecture of a compact student model with parallel branches which are trained using ground truth labels and information from high capacity teacher networks in an ensemble learning fashion. Our framework provides two main benefits: i) Distilling knowledge from different teachers into the student network promotes heterogeneity in  learning features at different branches of the student network and enables the network to learn diverse solutions to the target problem. ii) Coupling the branches of the student network through ensembling encourages collaboration and improves the quality of the final predictions by reducing variance in the network outputs.
Experiments on the well established CIFAR-10 and CIFAR-100 datasets show that our Ensemble Knowledge Distillation (EKD) improves classification accuracy and model generalization especially in situations with limited training data. Experiments also show that our EKD based compact networks outperform in terms of mean accuracy on the test datasets compared to other knowledge distillation based methods.
\end{abstract}
\section{Introduction}
Ensemble methods have shown considerable improvements in model generalization and produced state of the art results in several machine learning competitions (e.g., Kaggle) \cite{chen2017deeplab}. These ensemble methods typically contain multiple deep Convolutional Neural Networks (CNN) as sub-networks which are pre-trained on large-scale datasets to extract discriminative features from the input data. 
The size of an ensemble is not constrained by training because the sub-networks can be trained independently, and their outputs can be computed in parallel. However, in many applications limited training data is not sufficient to effectively train deep CNN models compared to small or compact networks. For instance in healthcare applications, the amount of available data is constrained by the number of patients. Therefore, improving generalization capability of compact network without requiring large-scale annotated datasets is of utmost importance.
Furthermore, today's high performing deep CNN based ensemble models have Giga-FLOPS compute and Giga-Bytes storage requirements \cite{huang2018gpipe}, making them prohibitive in resource constrained systems (e.g., mobile- or edge-devices) which have stringent requirements on memory, latency and computational power.
\\
\indent
To overcome these challenges, model compression techniques such as parameter pruning \cite{yu2018nisp} is a common way to reduce model size with trade-offs between accuracy and efficiency. Other techniques include hand crafting efficient CNN architectures such as SqueezeNets \cite{iandola2016squeezenet}, MobileNets \cite{howard2017mobilenets}, and ShuffleNets \cite{zhang2018shufflenet}. Recently, neural network search showed an effective way to generate efficient CNN architectures \cite{tan2019mnasnet,cai2018proxylessnas} by extensively tuning parameters such as network width, depth, filter types and sizes. These models showed better efficiency than hand-crafted networks but, at the cost of extremely large tuning cost. 
Another stream of work in building efficient networks for resource constrained scenarios is through Knowledge distillation \cite{hinton2015distilling}. It enables small low memory footprint networks to mimic the behavior of large complex networks by training small networks using the predictions of large networks as soft labels in addition to the ground truth hard labels. 
\\
\indent
In this paper we also explore Knowledge Distillation (KD) based strategies to improve model generalization and classification performance for applications with memory and compute restrictions. For this, we present a CNN architecture with parallel branches which distill high level features from different teacher networks during training and maintains low computational overhead during inference.
Our architecture provides two main benefits: \textbf{i)} It combines a student network with different teacher networks and distills diverse feature representations into the student network during training. This promotes heterogeneity in feature learning and enables the student network to mimic diverse high-level feature spaces produced by the teacher networks. \textbf{ii)} It combines the distilled information through parallel-branches in an ensembling manner. This reduces variance in the branch-level outputs and improves the quality of the final predictions of the student network.
In summary, the main contributions of this paper are as follows:
\begin{enumerate}
\item
We present an Ensemble Knowledge Distillation (EKD) framework which improves classification performance and model generalization of small and compact networks by  distilling knowledge from multiple teacher networks into a compact student network using an ensemble architecture.
\item 
We present a novel training objective function to distill ensemble knowledge into a single student network. Our objective function optimizes the parameters of the student network with a goal of learning mappings between input data and ground truth labels, and a goal of minimizing the difference between high level features of the teacher networks and the student network.  
\item 
We perform ablation study of our framework on CIFAR-10 and CIFAR-100 datasets in terms of different CNN architectures, varying ensemble sizes, and limited training data scenarios. Experiments show that by encouraging heterogeneity in feature learning through the proposed ensemble distillation, our EKD-based compact networks produce superior accuracy compared to the networks without using knowledge distillation. 
\end{enumerate}
\section{Related Work}
In this section, we discuss related work on model compression and knowledge distillation.
\subsection{Model Compression}
Network pruning is a popular approach to reduce a heavy network to obtain a light-weight form by removing redundancy in the heavy network. In this approach, a complex over-parameterized network is first trained, then pruned based on come criterions, and finally fine-tuned to achieve comparable performance with reduced parameters. In this context, methods such as \cite{yu2018nisp} compress large networks through the reduction of connections based on weight magnitudes or importance scores. Other methods used quantization of the weights to 8 bits \cite{han2015deep}, filter pruning \cite{li2016pruning} and channel pruning \cite{luo2017thinet} to reduce network sizes. However, the trimmed models are generally sub-graphs of the original networks and there is less flexibility in changing the original architecture design.
\subsection{Knowledge Distillation}
Knowledge Distillation (KD) aims at learning a light-weight student network such that it can mimic the behavior of a complicated teacher network. In this context, the work of \cite{ba2014deep} was the first to introduce knowledge distillation by minimizing L2 distance between the features from the last layers of two networks. Later, the work of \cite{hinton2015distilling} showed that the predicted class probabilities from the teacher are informative for the student and the probabilities can be used as a supervision signal in addition to the regular labeled training data during training. 
Romero et. al. \cite{romero2014fitnets} bridged the intermediate layers of the student and teacher networks in addition to the class probabilities and used L2 loss to supervise the student network. The method of \cite{czarnecki2017sobolev} minimized the difference between teacher and student derivatives of the loss combined with the divergence from the teacher predictions. Other methods explored knowledge distillation using activation maps \cite{heo2019knowledge}, attention maps \cite{zagoruyko2016paying}, Jacobians \cite{srinivas2018knowledge}, and unsupervised feature factors \cite{kim2018paraphrasing}.
\\
\indent
Ensembling is a promising technique to improve model generalization compared to the performance of individual models. Since different CNN architectures can achieve diverse distributions of errors due to the presence of several local minima, the combination of the outputs of individually trained networks leads to improved performance and better generalization to unseen test data. 
In the light of these studies, methods such as \cite{urban2016deep,furlanello2018born} combined ensemble learning and knowledge distillation to improve model generalization. For instance, the method of \cite{urban2016deep} trained an ensemble of 16 CNN models and compressed the learned function into shallow multi-layer perceptrons containing 5 layers.
The work of \cite{furlanello2018born} presented an iterative technique to transform a student model into the teacher model at each iteration. At the end of the iterations, all the student outputs were combined to form an ensemble. Our work also follows ensemble learning coupled with knowledge distillation however, compared to \cite{furlanello2018born}, we train a compact student network through knowledge distillation in a single iteration. Furthermore, our ensemble architecture distills knowledge from different teacher networks into the student network. 
This increases heterogeneity in student feature learning and enables the student network to mimic diverse feature representations produced by different teacher networks. Consequently, our EKD-based compact networks demonstrate better generalization capability compared to conventional KD methods \cite{hinton2015distilling}.
\section{The Proposed Framework}\label{proposed_framework}
\begin{figure*}[t!]
  \begin{center}
    \includegraphics[trim=0.0cm 0.2cm 0.0cm 0.0cm,clip,width=1.0\linewidth,keepaspectratio]{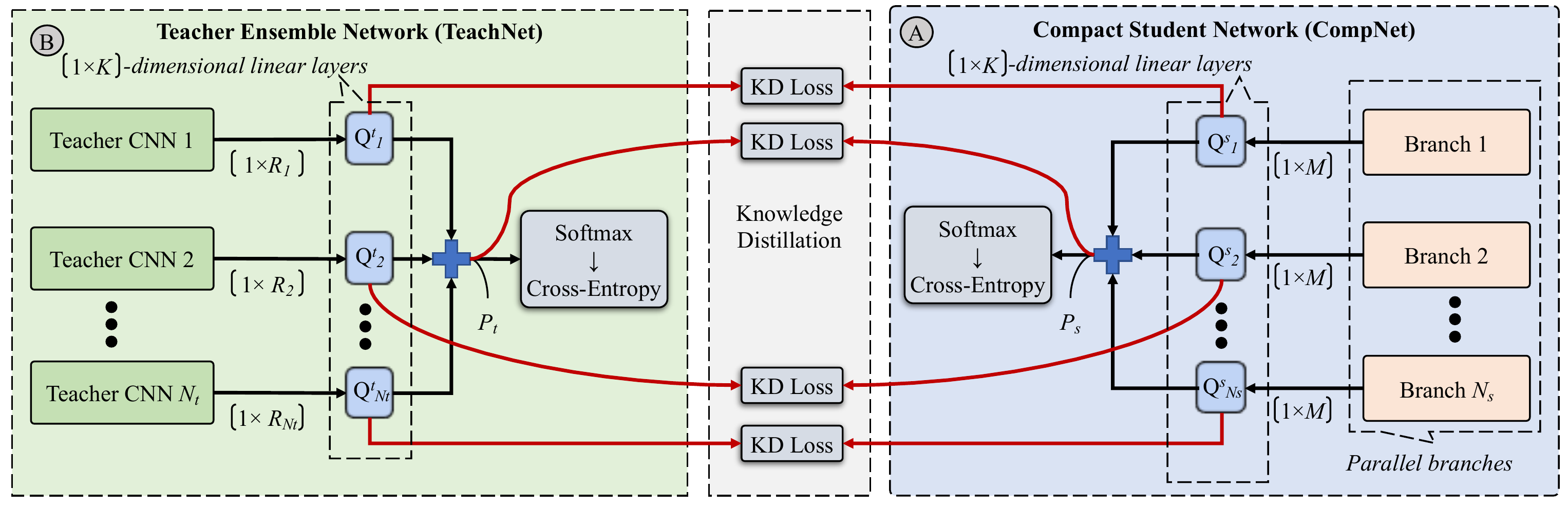}
  \vspace{-10pt}
  \caption{Overview of our framework which consists of a Compact Network (CompNet)-A and a teacher ensemble network (TeachNet)-B. CompNet is composed of parallel branches with similar architecture topology. During training, the branches are coupled with the sub-networks of the teacher ensemble and the parameters of the CompNet model are optimized with respect to the ground truth labels as well as the high-level features produced by the teacher ensemble. During testing, the branches of CompNet are executed in parallel (to increase inference speed) and their outputs are summed before Softmax to produce final predictions.} 
  \vspace{-5pt}
  \label{fig_network}
  \end{center}
\end{figure*}
Fig. \ref{fig_network} shows the overall architecture of our Ensemble Knowledge Distillation (EKD) framework. It consists of two main modules: \textbf{i)} A compact student network (CompNet) which is composed of $N_s$ branches connected in parallel (Fig. \ref{fig_network}-A). The branches follow a common architecture constituting convolutional and pooling layers. \textbf{ii)} A Teacher Ensemble Network (TeachNet) which is composed of $N_{t}$ CNN models with different architectures or layer configurations (Fig. \ref{fig_network}-B). In the following, we describe in detail the individual modules of the proposed framework.
\subsection{The Proposed Compact Network (CompNet)}\label{student_network}
Our compact network is composed of $N_s$ branches connected in parallel. The branches follow a common architecture where each branch is composed of multiple convolutional layers interconnected through residual connections. The branch outputs are fed into linear layers $(Q^s)$ to produce probabilistic distributions of the input data with respect to the target classes.
Our branch architecture is composed of ResNet structure which contain multiple residual blocks.
Specifically, the branch architecture starts with a $3\times3$ convolution followed by Batch Normalization (BN), and a Rectified Linear Unit (ReLU). Next, there are three residual blocks, where each residual block consists of two convolution layers with skip connections, followed by a pooling layer. Each convolution in the residual block is followed by BN, and a ReLU operation.
Each branch ends with global average pooling and produces $Y\in \mathbb{R}^{1\times M}-$dimensional feature maps which are then fed to a linear layer of $1\times K$ dimensions to produce probabilitic distributions ($Q_{s}\in\mathbb{R}^{1\times K}$) with respect to $K$ target classes. Mathematically, the output of a linear layer can be written as:
\begin{eqnarray}
Q^{s} = Y*W^s+B^s
\end{eqnarray}
where, $W^s$ and $B^s$ represent weights and bias matrices, respectively. Finally, the outputs of the linear layers are summed to produce a combined feature representation $P_s$ as shown in Fig. \ref{fig_network}-A. It is given by:
\begin{equation}\label{loss_cls}
P_s=\sum_{i=1}^{N_s}Q^{s}_i
\end{equation}
\subsection{The Proposed Teacher Ensemble Network (TeachNet)}\label{teacher_network}
Our teacher ensemble is composed of multiple CNN models which act as independent classifiers. The teacher sub-networks should use different architectures or layer configurations in order to produce diverse feature representations at the final convolutional layers. 
Similar to our CompNet architecture, the teacher outputs are first fed into linear layers to produce probabilistic distributions ($Q^{t}$) of the input data with respect to the target classes, and finally summed together to produce a combined feature representation $P_t$ as shown in Fig. \ref{fig_network}-B. It is given by:
\begin{equation}\label{loss_cls}
P_t=\sum_{i=1}^{N_t}Q^{t}_i
\end{equation}
In the following we describe our specialized training objective function which optimizes the parameters of our CompNet using ground truth labels as well as high-level feature representations from the teacher ensemble.
\subsection{The Proposed Ensemble Knowledge Distillation (EKD)}\label{deep_network}
Consider a training dataset of images and labels $(x,y)\in (\mathcal{X}, \mathcal{Y})$, where each sample belongs to one of the $K$ classes $(\mathcal{Y}={1,2,...,K})$. 
To learn the mapping $f_{s}(x): \mathcal{X}\rightarrow \mathcal{Y}$, we train our CompNet parameterized by $f_{s}(x,\theta^*)$, where $\theta^*$ are the learned parameters obtained by minimizing a training objective function $\mathcal{L}_{train}$:
\begin{eqnarray}\label{eq_seg} 
\theta^{*}=\argmin_{\theta}\mathcal{L}_{train}(y,f_{s}(x,\theta))
\end{eqnarray}
Our training function $\mathcal{L}_{train}$ is a weighted combination of three loss terms. A CrossEntropy loss term $\mathcal{L}_{CE}$ which is applied on the outputs of the teacher ensemble and the CompNet model with respect to the ground truth labels ($y$), and a distillation loss term $\mathcal{L}_{KD}$ which matches the outputs of the sub-networks of the teacher ensemble and the outputs of the branches of the CompNet model. Mathematically, $\mathcal{L}_{train}$ can be written as:
\begin{multline}\label{eq_train} 
\mathcal{L}_{train}=\alpha\cdot \mathcal{L}_{CE}(P_{t},y)+\beta\cdot \mathcal{L}_{CE}(P_{s},y)+\gamma\cdot\mathcal{L}_{KD},
\end{multline}
where $P_{t}=f_{t}(x)$ and $P_{s}=f_{s}(x)$ represent the logits (the inputs to the SoftMax) of the teacher ensemble and the CompNet model, respectively. The terms $\alpha \in [0,0.5,1]$, $\beta \in [0,0.5,1]$, and $\gamma \in [0,0.5,1]$ are the hyper-parameters which balance the individual loss terms. Mathematically, the CrossEntropy loss $\mathcal{L}_{CE}$ can be written as:
\begin{eqnarray}\label{eq_seg} 
\mathcal{L}_{CE}(P_{t},y)=\sum_{k=1}^K\mathbb{I}(k=y)\log\sigma (P_{t},y),
\end{eqnarray}
where $\mathbb{I}$ is the indicator function and $\sigma$ is the SoftMax operation. It is given by: 
\begin{equation}
\sigma(z)=\frac{\exp(z)}{\sum_{k=1}^K \exp(z_k)}. 
\end{equation} 
Our KD-based loss $\mathcal{L}_{KD}$ in Eq. \ref{eq_train} is composed of Kullback-Leibler (KL) divergence loss ($\mathcal{L}_{KL}$) and Mean-Squared-Error loss ($\mathcal{L}_{MSE}$).
Mathematically, $\mathcal{L}_{KD}$ can be written as:
\begin{multline}\label{eq_kd} 
\mathcal{L}_{KD}=\mathcal{L}_{KL}(P_{s},P_{t}/T)+\mathcal{L}_{MSE}(P_{s},P_{t})+\\
\sum_{i=1}^{N_s}(\mathcal{L}_{KL}(Q^{s}_i,Q^{t}_{i}/T)+\mathcal{L}_{MSE}(Q^{s}_{i},Q^{t}_{i})),
\end{multline}
where $i$ indexes the sub-networks of the teacher ensemble and the branches of the CompNet model. The term $T$ in Eq. \ref{eq_kd} is a temperature hyper-parameter which controls the softening of the output of the teacher sub-networks. A higher value of $T$ produces a softer probability distribution over the target classes.
The KL divergence loss $\mathcal{L}_{KL}$ is defined between log-probabilities computed from the outputs of a student network and probabilities computed from the outputs of a teacher network. Mathematically, it can be written as: 
\begin{eqnarray}\label{eq_seg} 
\mathcal{L}_{KL}(Q^{s}_1,Q^{t}_1)=\sum_{k=1}^K \sigma(Q^{s}_1)\cdot\left(\log(\sigma(Q^{s}_1)) - \sigma(Q^{t}_1) \right),
\end{eqnarray}
where $\sigma$ represents the SoftMax operation.
\section{Experimental Setup}\label{implementation}
In this section we describe the details of our experiments.
\subsection{Network Architectures}
\begin{table*}[t!]
	\caption{Ablation study of our ensemble distillation based student network in terms of the Ensemble Size (ES), number of training parameters, number of FLOPS, and mean test accuracy on CIFAR-10, CIFAR-100, and Tiny ImageNet datasets for ResNet8 as the student network architecture.}
	\vspace{-0pt}
	\centering
	\setlength\tabcolsep{5.0pt}\centering
	\begin{tabular}{@{}cccccccccccc@{}}
		\toprule
		\multirow{2}{*}{Model} &\multirow{3}{*}{ES}&\multirow{3}{*}{Teachers}&\multicolumn{2}{c}{CIFAR10}&\multicolumn{2}{c}{CIFAR100}&\multicolumn{2}{c}{Tiny ImageNet}&No. of param&No. of FLOPS\\
		&&&no EKD&with EKD&no EKD&with EKD&no EKD&with EKD&(million)&(million)\\
		\midrule
		\parbox[t]{15mm}{\multirow{7}{*}{\rotatebox[origin=c]{0}{ResNet8}}}
		& 1&T1&87.51&\textbf{89.66}			&60.03&\textbf{60.57}		&34.98&\textbf{36.84}&0.08&12.75\\	
		& 2&T1-T2&86.60&\textbf{91.11}	&60.58&\textbf{62.79}		&37.85&\textbf{40.36}&0.16&25.50\\	
		& 3&T1-T3&87.41&\textbf{91.40}	&62.78&\textbf{64.09}		&37.99&\textbf{41.58}&0.23&38.25\\	
		&4&T1-T4&87.74&\textbf{91.72}	&63.25&\textbf{65.45}	&38.00&\textbf{41.86}&0.31&51.01\\			
		& 5&T1-T5&87.81&\textbf{92.15}	&64.24&\textbf{66.54}	&38.16&\textbf{42.63}&0.39&63.76\\			
		&6&T1-T6&88.14&\textbf{92.24}	&60.75&\textbf{67.36}		&39.45&\textbf{42.59}&0.47&76.51\\			
		\rowcolor{grey1}
		& 7&T1-T7&88.05&\textbf{92.33}	&60.83&\textbf{67.78}&39.06&\textbf{43.89}&0.55&89.26\\	
		\bottomrule
	\end{tabular}
	\vspace{-0pt}
	\label{table_ensemble_size}
\end{table*}
\begin{figure*}[t!]
	\begin{center}
		\includegraphics[trim=0.1cm 0.1cm 0.1cm 0.1cm,clip,width=1.0\linewidth,keepaspectratio]{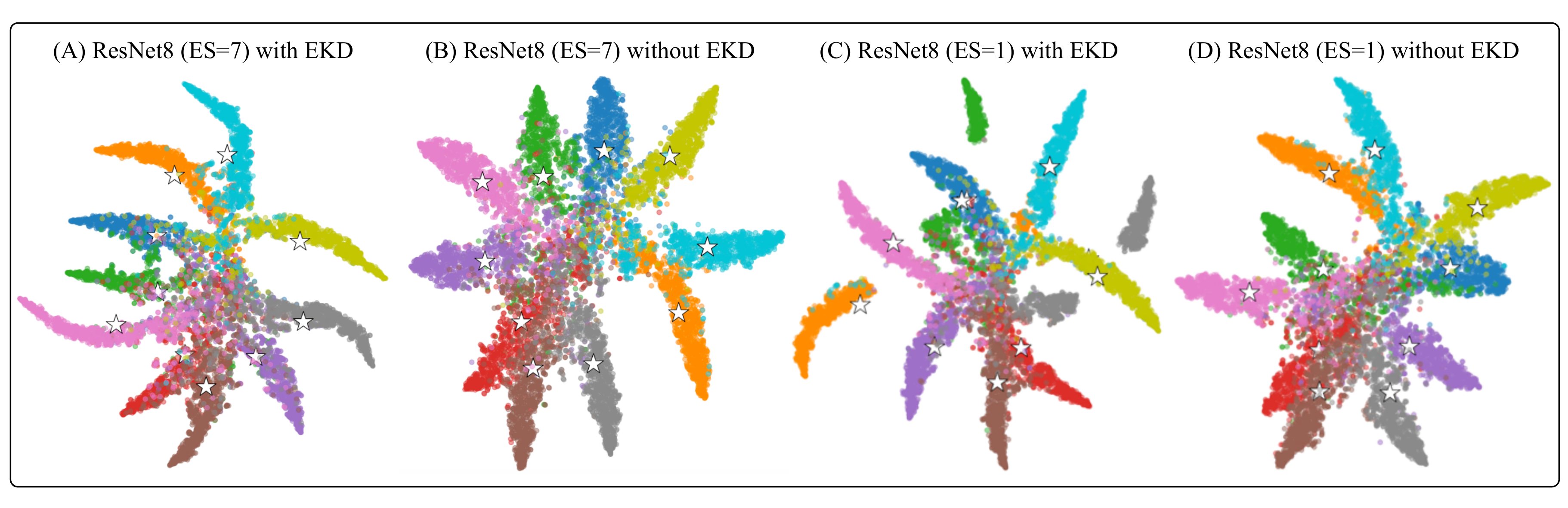}\\
		\vspace{-5pt}
		\caption{Comparison of TSNE visualizations of 2-dimensional embeddings generated using features produced by a 7-branch ResNet8 model with EKD (A), a 7-branch ResNet8 model without distillation (B), a 1-branch ResNet8 model with EKD (C), and a 1-branch ResNet8 model without distillation (D) on the test data of CIFAR-10 dataset. The comparison shows that the embeddings produced by our EKD based models show better separation of the target classes compared to the embeddings produced by models trained without distillation.}
		\vspace{-5pt}
		\label{fig_tsne1}
	\end{center}
\end{figure*}
We evaluated our ensemble knowledge distillation using ResNet8 structure as the student network as used in other knowledge distillation based studies \cite{heo2018improving,mirzadeh2019improved}. For the teacher ensemble, we considered up to 7 sub-networks based on ResNet14, ResNet20, ResNet26, ResNet32, ResNet44, ResNet56, and ResNet110 architectures as used in  \cite{heo2018improving,mirzadeh2019improved}.
\subsection{Training and Implementation}
For training, we initialized the weights of the convolutional and the fully connected layers from zero-mean Gaussian distributions. The standard deviations were set to 0.01, the biases were set to 0, and a parameter decay of 0.0005 was set on the weights and biases. The teacher ensemble was first trained independently from the scratch, and then fine-tuned simultaneously and collaboratively with the student network. The distillation from the teacher sub-networks to the student network was performed throughout the whole training process by optimizing the training objective function in Eq. \ref{eq_train}.     
Specifically, we trained the networks for 500 epochs starting with a learning rate of 0.01 which was divided by 10 at 50\% and 75\% of the total number of epochs. 
Our implementation is based on the auto-gradient computation framework of the Torch library \cite{paszke2017automatic}. Training was performed by ADAM optimizer with a batch size of 128 using 2 nvidia V100 GPU hardware.
For hyper-parameter optimization, we used the toolkit of \cite{bergstra2011algorithms} to tune the loss weighing parameters $\alpha=0.5, \beta=0.5, \gamma=0.6$, and the temperature parameter $T=10$. 
\subsection{Datasets}
We evaluated our framework on three well calibrated image classification datasets CIFAR-10, CIFAR-100, and Tiny ImageNet\footnote{https://tiny-imagenet.herokuapp.com/}. CIFAR-10 and CIFAR-100 consist of 60,000 RGB images distributed into 10 and 100 classes, respectively. Specifically, the training set contains 50,000 images and the test set contains 10,000 images of sizes $64\times64$ pixels.
Tiny ImageNet is a subset of the ImageNet dataset. It contains 100k training images and 10k test images distributed into 200 classes. 
\section{Results and Analysis}
\begin{figure*}[t!]
	\begin{center}
		\includegraphics[trim=0.1cm 0.1cm 0.1cm 0.1cm,clip,width=1.0\linewidth,keepaspectratio]{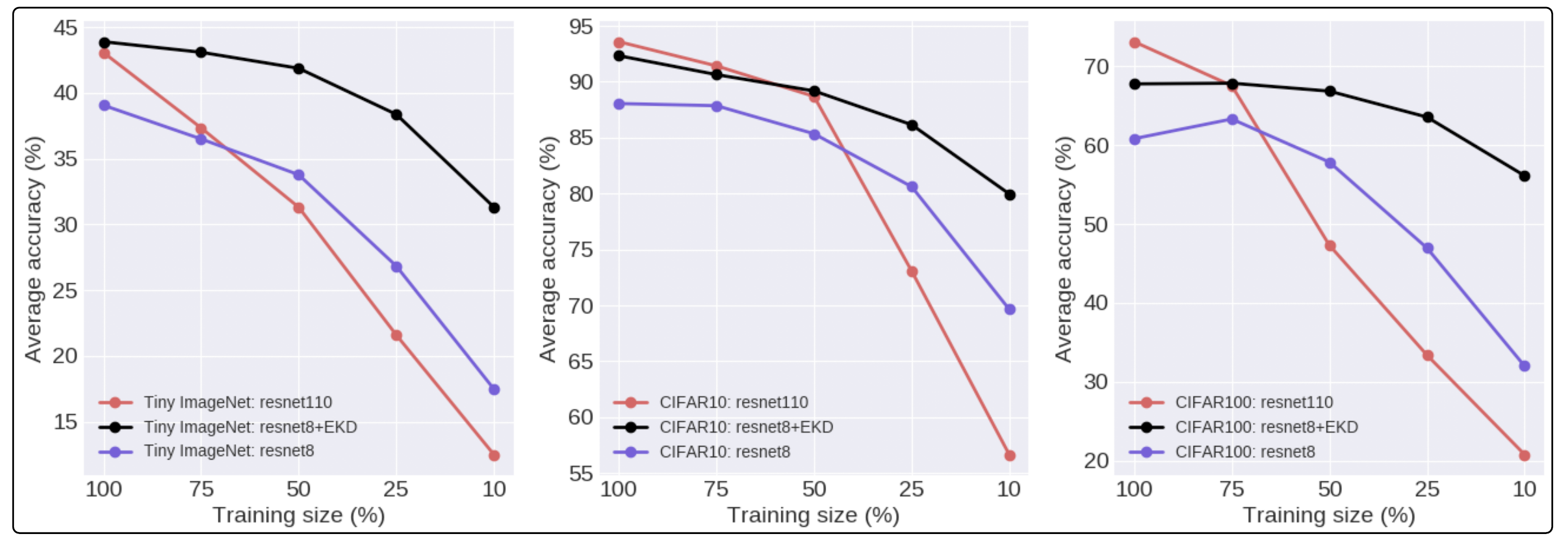}
		\vspace{-10pt}
		\caption{Comparison of mean accuracy of a 7-branch ResNet8 model with and without the proposed EKD for different sizes of data used for training the models on CIFAR-10, CIFAR-100, and Tiny ImageNet datasets. The comparison shows that compared to ResNet8 without using knowledge distillation, our EKD-based ResNet8 produced considerably higher mean accuracy for all the tested data sizes. Our EKD-based 7-branch ResNet8 also produced higher mean accuracy compared to ResNet110 when limited data was used to train the models.}
		\vspace{-10pt}
		\label{fig_resnet110}
	\end{center}
\end{figure*}
\begin{figure*}[t!]
	\begin{center}
		\includegraphics[trim=0.1cm 0.1cm 0.1cm 0.1cm,clip,width=1.0\linewidth,keepaspectratio]{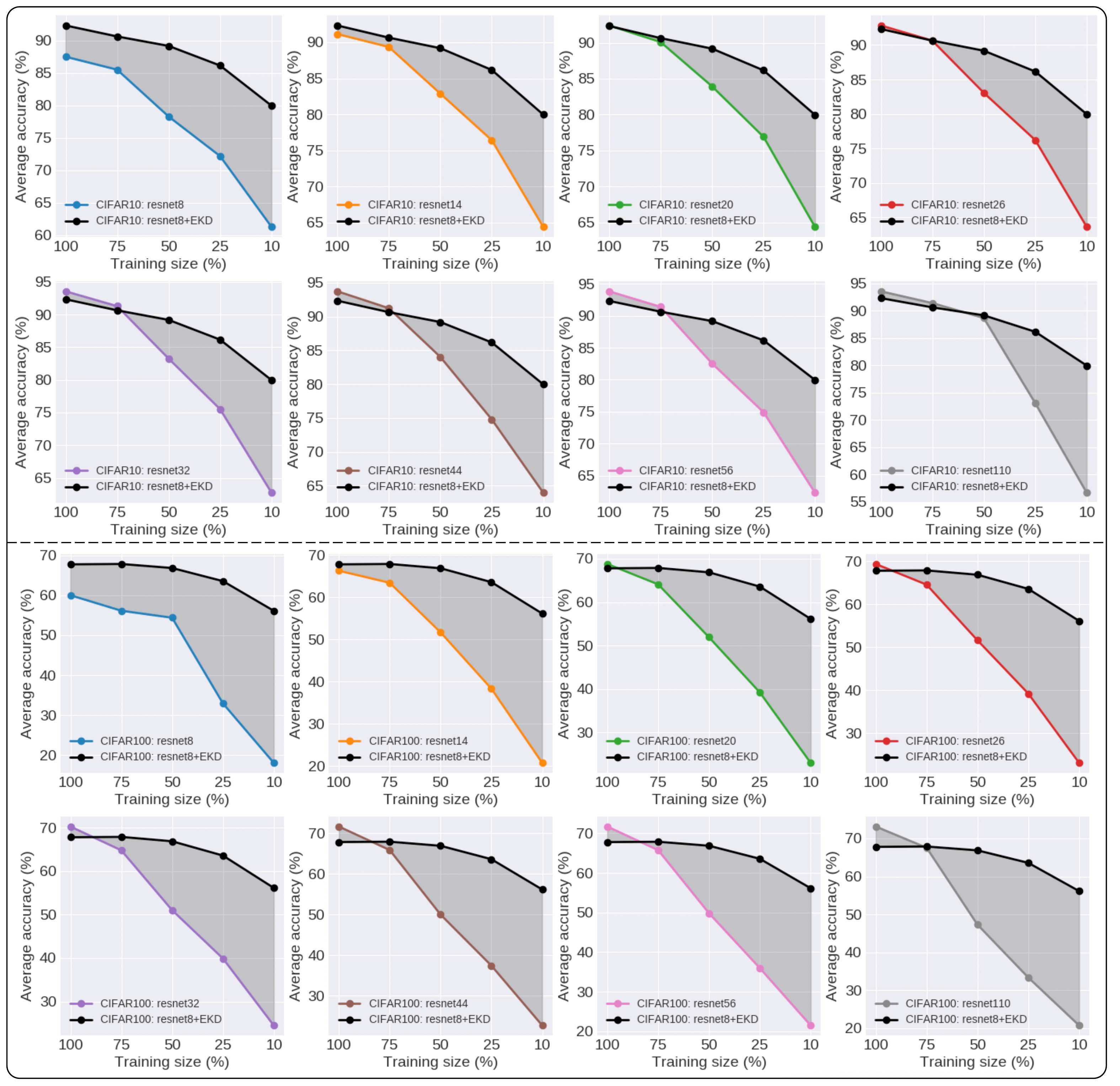}\\
		\vspace{0pt}
		\caption{Comparison of mean accuracy of our EKD-based ResNet8 model and teacher networks for different sizes of data used for training on CIFAR-10 and CIFAR-100 datasets. The comparison shows that the proposed EKD-based ResNet8 produces higher accuracy compared to the models trained without knowledge distillation under limited training data cases.}
		\vspace{-5pt}
		\label{fig_teacher_comp}
	\end{center}
\end{figure*}
\begin{table*}[t]
	\caption{Ablation study of a 7-branch ResNet8 model (ES=7) with and without the proposed EKD and the proposed teacher networks in terms of number of training parameters, number of FLOPS, and inference speed. The results show that the proposed EKD based ResNet8 produces higher average accuracy compared to the other networks without knowledge distillations on all the tested datasets.}
	\vspace{-0pt}
	\centering
	\setlength\tabcolsep{5pt}\centering
	\begin{tabular}{@{}lcccccc@{}}
		\toprule
		\multirow{2}{*}{Model}&CIFAR-10&CIFAR-100&Tiny ImageNet&No. of param.&No. of FLOPS&Inference time\\
		&data size 10\%&data size 10\%&data size 10\%&(million)&(million)&(ms)\\					
		\midrule
		ResNet14	&64.39&20.77&15.89	&0.19&27.09&2\\
		ResNet20	&64.38&22.98&16.67	&0.28&41.42&5\\
		ResNet26	&63.61&23.13&14.79	&0.38&55.75&5\\
		ResNet32	&62.74&24.48&13.69	&0.48&70.07&6\\
		ResNet44	&63.95&22.74&14.80	&0.67&98.73&9\\
		ResNet56	&62.33&21.48&16.14	&0.87&127.39&10\\
		ResNet110	&56.66&20.75&12.51&1.74&256.34&18\\
		\hline
		\rule{-2.5pt}{3ex}
		ResNet8 (ES=7)&69.65&31.98&17.49&0.55&89.26&11\\
		ResNet8+EKD (ES=7)&\textbf{79.97}&\textbf{56.13}&\textbf{31.32}&0.55&89.26&11\\
		\bottomrule
	\end{tabular}
	\vspace{-0pt}
	\label{table_teachers}
\end{table*}
\subsection{Ensemble Distillation Improves Model Performance}
Here, we evaluate our compact student networks with and without the proposed Ensemble Knowledge Distillation (EKD) on the CIFAR-10, CIFAR-100, and Tiny ImageNet datasets. Table \ref{table_ensemble_size} shows the results of these experiments for different ensemble sizes. From Table \ref{table_ensemble_size} we see that EKD based networks improve accuracy for all the tested ensemble sizes on all the target datasets. For instance, EKD with an ensemble size of 7 improves accuracy by upto 4\%, 7\%, and 4\% on the CIFAR-10, CIFAR-100, and Tiny ImageNet datasets, respectively. 
\subsection{Ensemble Learning Improves Knowledge Distillation}
Here, we evaluate the performance of our EKD based student networks by varying the size of the ensemble to explore the benefits of ensembling for knowledge distillation. Table \ref{table_ensemble_size} shows the results of these experiments on CIFAR-10, CIFAR-100, and Tiny ImageNet datasets. The results show that the accuracy increases with the increase in ensemble size (ES) at the cost of increase in the number of parameters and the number of FLOPS. For instance, a student network with 7 branches improves accuracy by around 2\%, 7\%, and 7\% compared to a 1-branch student network on the CIFAR-10, CIFAR-100, and Tiny ImageNet datasets, respectively. Fig. \ref{fig_tsne1} shows a comparison of TSNE embeddings of features produced by ResNet8 models with different ensemble sizes. From the figure, it is clear that the embeddings produced by the 7-branch model with EKD shows better seperation of the target classes compared to the embeddings produced by the 1-branch model without distillation.
\subsection{Model Generalization Performance}
Here we evaluate the generalization performance of the proposed ensemble knowledge distillation. For this, we conducted experiments for different sizes of the data used for training the teacher and the student networks. Fig. \ref{fig_resnet110} and Fig. \ref{fig_teacher_comp} show the results of these experiments using ResNet8 as the student network. The results show that the performance gap increases for all the tested networks as the size of the dataset is reduced. For instance, the accuracy drop by around 25\%, 40\%, and 30\% when 10\% of the data was used to train the networks without the proposed EKD on CIFAR-10, CIFAR-100, and Tiny ImageNet datasets, respectively as shown in Fig. \ref{fig_resnet110}.
Fig. \ref{fig_resnet110} and Fig. \ref{fig_teacher_comp} also show that using the proposed EKD, ResNet8 based models improve test accuracy for all the tested sizes of training data with considerable margins compared to ResNet8 models without distillation and the other networks used as teachers (ResNet14, ResNet20, ResNet26, ResNet32, ResNet44, ResNet56, and ResNet110). 
For instance, EKD-based ResNet8 models produced improvements of upto 12\%, 10\%, and 25\% when 10\% of data was used for training on CIFAR-10, CIFAR-100 and Tiny ImageNet datasets, respectively compared to networks without distillation as shown in Fig. \ref{fig_resnet110}. 
These improvements are attributed to our ensemble distillation which promotes diversity in feature learning by transfering knowledge from different teachers into the student network and improves model generalization to test data.
These experiments show that for situations where non-KD methods fail to achieve generalization due to insufficient data, the proposed ensemble distillation achieves considerable performance improvements, thereby demonstrating potentials for uses in applications with limited-data constraints.
\\
\indent
Table \ref{table_teachers} shows a comparison between our EDK-based 7-branch student network and the teacher networks on the CIFAR-10, CIFAR-100, and Tiny ImageNet datasets when 10\% of the data was used for training. The results show that our EKD based 7-branch ResNet8 produced the best accuracy on the test datasets with $3\times$ less number of parameters, $2.8\times$ less number of FLOPS, and faster inference speed compared to the ResNet110 network.
\begin{figure*}[t!]
	\begin{center}
		\includegraphics[trim=0.1cm 0.1cm 0.1cm 0.1cm,clip,width=1.0\linewidth,keepaspectratio]{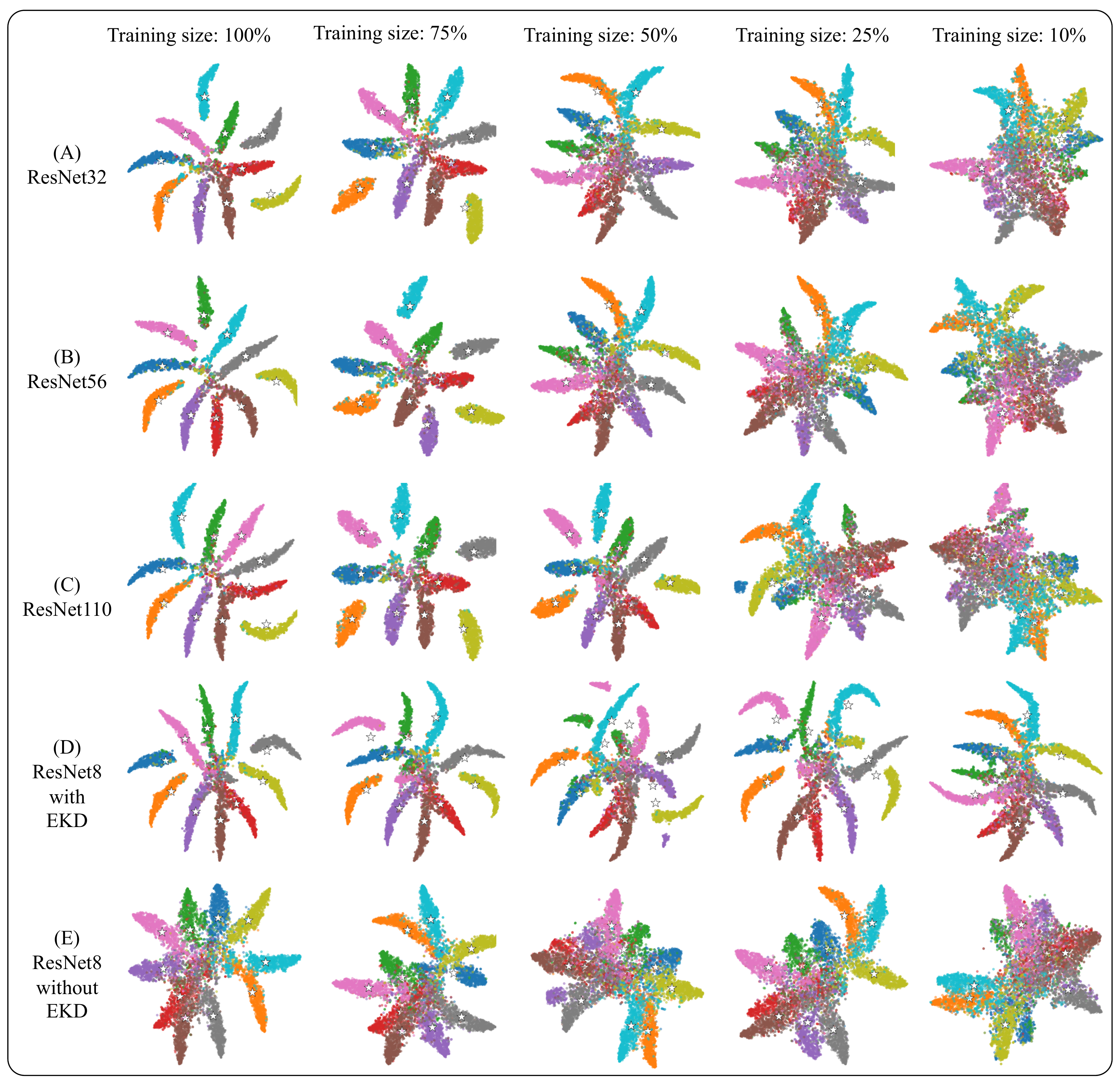}
		\vspace{-15pt}
		\caption{Comparison of TSNE visualizations of 2-dimensional embeddings generated using features produced by the proposed teacher networks (A, B, and C), our EKD-based ResNet8 (D), and a ResNet8 model without distillation (E), on the test data of CIFAR-10 dataset. The comparison shows that the embeddings produced by our EKD based models show better separation of the target classes especially in cases with limited training data compared to the embeddings produced by models trained without distillation.}
		\vspace{-5pt}
		\label{fig_tsne}
	\end{center}
\end{figure*}
We also conducted experiments to visualize the features space learnt by our EKD based networks. Fig. \ref{fig_tsne} shows the 2-dimensional TSNE-embeddings generated using the features produced by the teacher networks and our 7-branch ResNet8 models with and without using the proposed ensemble distillation under different sizes of data used during training. These experiments show that training a student network using high-level features from multiple teacher networks enables the student network to produce embeddings which exhibit better separation of the target classes compared to the embeddings produced by the models without using distillation.
\subsection{Comparison with Other Knowledge Distillation Methods}
\begin{table}
	\caption{Comparison of our ensemble distillation using 1-branch ResNet8 base student network and other Knowledge Distillation (KD) methods on the CIFAR-10 dataset.}
	\vspace{-0pt}
	\centering
	\setlength\tabcolsep{6.0pt}\centering
	\begin{tabular}{@{}lccccccc@{}}
		\toprule
		\multirow{1}{*}{Method}&\multirow{1}{*}{Dataset} &\multicolumn{1}{c}{Accuracy}\\
		\midrule
		Hinton \cite{hinton2015distilling}& CIFAR-10&86.66\\
		FITNET \cite{mirzadeh2019improved}&CIFAR-10&86.73\\
		Attention \cite{zagoruyko2016paying} &CIFAR-10&86.86\\
		FSP \cite{yim2017gift} &CIFAR-10&87.07\\
		BSS \cite{heo2018improving} &CIFAR-10&87.32\\
		MUTUAL \cite{zhang2018deep} & CIFAR-10&87.71\\
		TAKD \cite{mirzadeh2019improved} &CIFAR-10&88.01\\ 
		(this work) ResNet8+EKD (ES=1)& CIFAR-10&\textbf{89.66}\\					
		\bottomrule
	\end{tabular}
	\vspace{-0pt}
	\label{table_sota}
\end{table}
Here, we compare our Ensemble Knowledge Distillation (EKD) with some of the recent state-of-the-art knowledge distillation based methods including: activation based attention transfer (AAT) \cite{zagoruyko2016paying}, the method of \cite{yim2017gift} (FSP), the method of \cite{hinton2015distilling}, hint based transfer (FitNet) \cite{romero2014fitnets}, the method of \cite{heo2018improving}(BSS), the method of deep mutual learning \cite{zhang2018deep}(MUTUAL), and the method of \cite{mirzadeh2019improved} (TAKD). For a fair comparison, we used exactly the same setting for CIFAR-10 experiments, and a ResNet8 based student network as used in the baseline studies. Table \ref{table_sota} shows that our EKD based ResNet8 improved accuracy on the tested dataset compared to the other KD methods.
This improved performance is attributed to the proposed ensemble distillation architecture where the proposed training objective function enabled the student network to successfully mimic diverse feature embeddings produced by different teachers and improve generalization to unseen test data. Furthermore, the combination of distilled information through ensembling reduced variance in the outputs and improved the quality of the final predictions of the student network.
\subsection{Conclusion and Future Work}
Recently, deep CNN based ensemble methods have shown state-of-the-art performance in image classification but at the cost of high computation cost and large memory requirements.
In this paper, we show that knowledge distillation using an ensemble architecture can improve classification accuracy and model generalization especially with fewer training data for small and compact networks.
Unlike traditional ensembling techniques which reduce variance in outputs by combining independently trained networks, we show that Ensemble Knowledge Distillation (EKD) encourages heterogeneity in student feature learning through collaboration between different teachers and the student network. This enables student networks to learn more discriminative and diverse feature representations while maintaining small memory and compute requirements.  
Experiments on well-established CIFAR-10, CIFAR-100, and Tiny ImageNet datasets show that compact networks trained through the proposed ensemble distillation improved classification accuracy and model generalization especially in situations with fewer training data.
In future, we plan to explore a fully data-driven automated ensemble selection. We also plan to evaluate our framework for video classification tasks to gain more insights into the benefits of ensemble distillation.    
\bibliographystyle{ecai}
\bibliography{ijcai19}
\end{document}